\documentclass[conference]{IEEEtran}
\IEEEoverridecommandlockouts
\usepackage{cite}
\usepackage{amsmath,amssymb,amsfonts}
\usepackage{algorithmic}
\usepackage{graphicx}
\usepackage{textcomp}
\usepackage{xcolor}
\usepackage{times}
\usepackage{soul}
\usepackage{url}
\usepackage[hidelinks]{hyperref}
\usepackage[utf8]{inputenc}
\usepackage[small]{caption}
\usepackage{amsmath}
\usepackage{booktabs}
\usepackage{algorithm}
\usepackage{algorithmic}
\usepackage{subfigure}
\usepackage{latexsym}
\usepackage{algorithm}
\usepackage{algorithmic}
\usepackage{url}
\usepackage{multirow}
\def\BibTeX{{\rm B\kern-.05em{\sc i\kern-.025em b}\kern-.08em
    T\kern-.1667em\lower.7ex\hbox{E}\kern-.125emX}}
\begin{document}

\title{Multi-source Transfer Learning with Ensemble for Financial Time Series Forecasting
}

\author{\IEEEauthorblockN{1\textsuperscript{st} Qi-Qiao He}
\IEEEauthorblockA{\textit{Department of Computer and Information Science} \\
\textit{University of Macau}\\
Macau, China\\
yc07422@umac.mo}
\and
\IEEEauthorblockN{2\textsuperscript{nd} Patrick Cheong-Iao Pang}
\IEEEauthorblockA{\textit{Victoria University Business School} \\
\textit{Victoria University}\\
Melbourne, Australia \\
mail@patrickpang.net}
\and
\IEEEauthorblockN{3\textsuperscript{rd} Yain-Whar Si}
\IEEEauthorblockA{\textit{Department of Computer and Information Science} \\
\textit{University of Macau}\\
Macau, China \\
fstasp@umac.mo}
}

\maketitle
\begin{abstract}
	Although transfer learning is proven to be effective in computer vision and natural language processing  applications, it is rarely investigated  in forecasting financial time series. Majority of existing works on transfer learning are based on single-source transfer learning due to the availability of open-access large-scale datasets. However, in financial domain, the lengths of individual time series are relatively short and single-source transfer learning models are less effective. Therefore, in this paper, we investigate multi-source deep transfer learning for financial time series. We propose two multi-source transfer learning methods namely Weighted Average Ensemble for Transfer Learning (WAETL) and Tree-structured Parzen Estimator Ensemble Selection (TPEES).  The effectiveness of our approach is evaluated on financial time series extracted from stock markets. Experiment results reveal that TPEES outperforms other baseline methods on majority of multi-source transfer tasks.
\end{abstract}


\section{Introduction}


Time series forecasting is one of the challenging research problems in financial domain.  
However, majority of transfer learning researches for time series focus on single-source transfer learning, meaning that only a single source dataset is used for training the models. However, when compared to image and text datasets used in training deep learning models for Computer Version (CV) and Natural Language Processing (NLP) applications, a single time series data (e.g. historical price data of a listed company from stock markets) is relatively small (short). In these situations, training process could result in overfitting models. In order to alleviate this problem, we investigate multi-source transfer learning models for forecasting financial time series in this paper. One of the key factors in adopting multiple data sources for transfer learning is motivated by the fact that the future price of a stock could be influenced by the historical prices of stocks within the same industry/sector.  For example, the future price trend of Hongkong and Shanghai Banking Corporation (HSBC) could be correlated to the prices of other banks in Hong Kong and Asia Pacific region. In this paper, we aim to exploit this correlation property for generating better deep learning models. 
In addition, in the context of time series forecasting, the features of time series and the calculation of similarity between two time series are inherently different from CV and Natural Language Processing applications. Besides, in transfer learning for time series forecasting, existing algorithms rarely exploit the similarity between two time series. 

Against this background, in this paper, we propose two ensemble methods for multi-source transfer learning. They are both parameter-based transfer learning methods~\cite{karl2016survey}. The proposed ensemble methods combine multiple models, each of which is pre-trained by a different source dataset and fine-tuned by the same target dataset. In the first ensemble method called Weighted Average Ensemble for Transfer Learning (WAETL), weights are calculated based on the similarity between source and target time series datasets. In WAETL, models with poor performance are assigned with smaller weights than good performance models. 
Extensive experiments are also conducted to investigate the effect of distance functions on the transferred models. The second method called Tree-structured Parzen Estimator Ensemble Selection (TPEES) is based on Tree-structured Parzen Estimator (TPE) optimization. In this approach, we treat the process of selecting models from transfer learning model pool as an optimization problem. WAETL use all models in model pool, but some models in model pool may not be selected by TPEES.
The contributions of this paper can be summarized as follows:

\begin{itemize}
	\item Two novel ensemble based multi-source transfer learning methods called WAETL and TPEES are proposed for financial time series forecasting. The proposed approaches aim to alleviate the problem of insufficient training data when forecasting stock prices in financial markets.
	\item Extensive analysis are also performed to investigate the effect of different distance functions to calculate the similarity of time series. Experiment results shows that WD and Coral achieve best results when they are applied with WAETL method.
\end{itemize}
The rest of the paper is structured as follows. In section 2, we review existing work on multi-source transfer learning for financial time series forecasting. In section 3, we describe our proposed methods. In section 4, we present our experiment results. Finally, we conclude the paper with future work in section 5.

\section{Background and Related Work}

In~\cite{Ding}, Ding et al. combined the neural tensor network and deep CNN to predict the short--term and long--term influences of events on stock price movements. A deep learning framework based on long-short term memory (LSTM) was also proposed by Bao et al.~\cite{bao2017deep} for time series forecasting. However, when the available data is insufficient for training, the performance of deep learning model can be poorer than traditional statistical methods~\cite{Makridakis}. Besides, training a deep learning model can be time-consuming and expensive. In order to alleviate the above-mentioned problems, transfer learning has been combined with deep learning in~\cite{Yosinski}. 

Recently, transfer learning was adopted for analyzing time series data. Fawaz et al.~\cite{Fawaz} investigate how to transfer deep CNNs for Time Series Classification (TSC) tasks. Laptev et al. \cite{Laptev} also propose a new loss function and an architecture for time series transfer learning. Ye et al.~\cite{Ye} propose a novel transfer learning framework for time series forecasting. In these approaches, one source dataset is used for pre-training and one target dataset is used for fine-tuning, called single-source transfer learning. In this paper, we use single-source transfer learning as a baseline method. 


Multi-task learning (MTL) is a parameter based multi-source transfer learning method, which is successfully used in CV and NLP. The goal of MTL is to improve the performance of each individual task by leveraging useful information between multiple related learning tasks \cite{Huang}. In~\cite{hu2016transfer}, MTL is used to forecast short-term wind speed. In this paper, we adopt a model similar to MTL approach in which both all source and target datasets are used for training and target dataset is used to fine-tune the MTL model in the final step. Therefore, the MTL model shares all the hidden layers except the output layer. 

Christodoulidis et al.~\cite{christodoulidis2016multisource} transfers knowledge from multiple source datasets to a target model with ensemble method called forward ensemble selection (FES) to classify lung pattern. In their approach, CNN is used for classification. 
First, Christodoulidis et al. utilize improving ensemble selection procedure to select fine-tuned CNN models from model pool. Next, a simple average combination method is used to build an ensemble model. 
In this paper, we proposed two new ensemble methods for multi-source transfer learning to build a strong ensembled model. In addition, we adopt FES as a baseline method for comparison. However, in order to forecast time series, we replace CNN with LSTM and Multilayer Perceptrons (MLP).

\section{Methods}



Multi-source transfer learning with ensemble relaxed the assumption of MTL. When some of the models pre-trained by the source datasets and fine-tuned by the target dataset have a negative effect on the target model, multi-source transfer learning with ensemble can mitigate the impact of these models on the target model. Meanwhile, Multi-source transfer learning with ensemble focus on improving the performance of target task with multi-source datasets. Therefore, in this paper, we propose two ensemble methods for multi-source transfer learning namely Weighted Average Ensemble for Transfer Learning (WAETL) and Tree-structured Parzen Estimator Ensemble Selection (TPEES). Model pool contains fine-tuned models which have been pre-trained by source datasets and fine-tuned by the target dataset. We use above ensemble methods to combine the output of each model from the model pool. 
To evaluate their effectiveness, these two methods are compared with Average Ensemble (AE) and Forward Ensemble Selection (FES) methods in the experiments.

\subsection{Weighted Average Ensemble for Transfer Learning (WAETL)}
Averaging ensemble (AE) is one of the most common ensemble methods~\cite{tyagi2014survey}. The aggregated output of target model is averaged by the output of each model from model pool. Simple averaging avoids overfitting and creates smoother ensemble model. Therefore, AE is used as baseline method in this paper. However, not all models from model pool have same influence on the target model. Hence, Weighted Averaging Ensemble (WAE) based on Average Ensemble (AE) is proposed in~\cite{Touretzky1997Learning}. In this paper, we further extend WAE for multi-source transfer learning.

Unlike AE, where each fine-tuned model has the same weight, the proposed WAETL can increase the importance of one or more fine-tuned models. Rosenstein et al. \cite{Rosenstein} empirically showed that if the source and target datasets are dissimilar, then brute-force transfer may negatively effect the performance of the target dataset. Such effect is also labeled as negative transfer by \cite{Rosenstein}. Mignone et al.~\cite{mignone2019exploiting} compute a weight for each instance according to their similarity with clusters in source and target datasets, which is quite recognized. This method is instance-based transfer learning approach and focus on single source transfer learning. However, WAETL is parameter-based transfer learning approach and compute a weight for each source dataset according to their similarity with the target dataset.
In WAETL, we use different distance functions including CORrelation ALignment (CORAL) loss \cite{sun2016return}, Wasserstein Distance (WD) \cite{Ruschendorf1985Wasserstein}, Dynamic Time Warping (DTW) \cite{Berndt1994Dynamic}, Pearson Correlation Coefficient (PCC) \cite{benesty2009pearson} to calculate the similarity between each source and target domain. The similarity value $D(s_i, t)$ calculated by above distance functions can be used as weight $w_i$ in WAETL through a function $f(D(s_i, t))$. The larger the weight $w_i$, the more influence the corresponding $i^{th}$ model has on the target model. The output ($out$) of target model can be formulated as Equation \ref{equation: waetl1} and \ref{equation: waetl2}.

\begin{equation}\small
w_i = f(D(s_i, t)),
\label{equation: waetl1}
\end{equation}
\begin{equation}\small
out = \sum\limits_{i=1}^{n} w_i * out_i, \quad \text{where} \sum\limits_{i=1}^{n} w_i = 1
\label{equation: waetl2}
\end{equation}
where  $out_i$ is output of the $i^{th}$ model in the model pool and $n$ is the size of model pool. $s_i$ is the $i^{th}$ source dataset and $t$ is the target dataset.

\subsection{Tree-structured Parzen Estimator Ensemble Selection (TPEES)}
\begin{figure}[!htbp]
	\centering
	\includegraphics[width=0.4\textwidth]{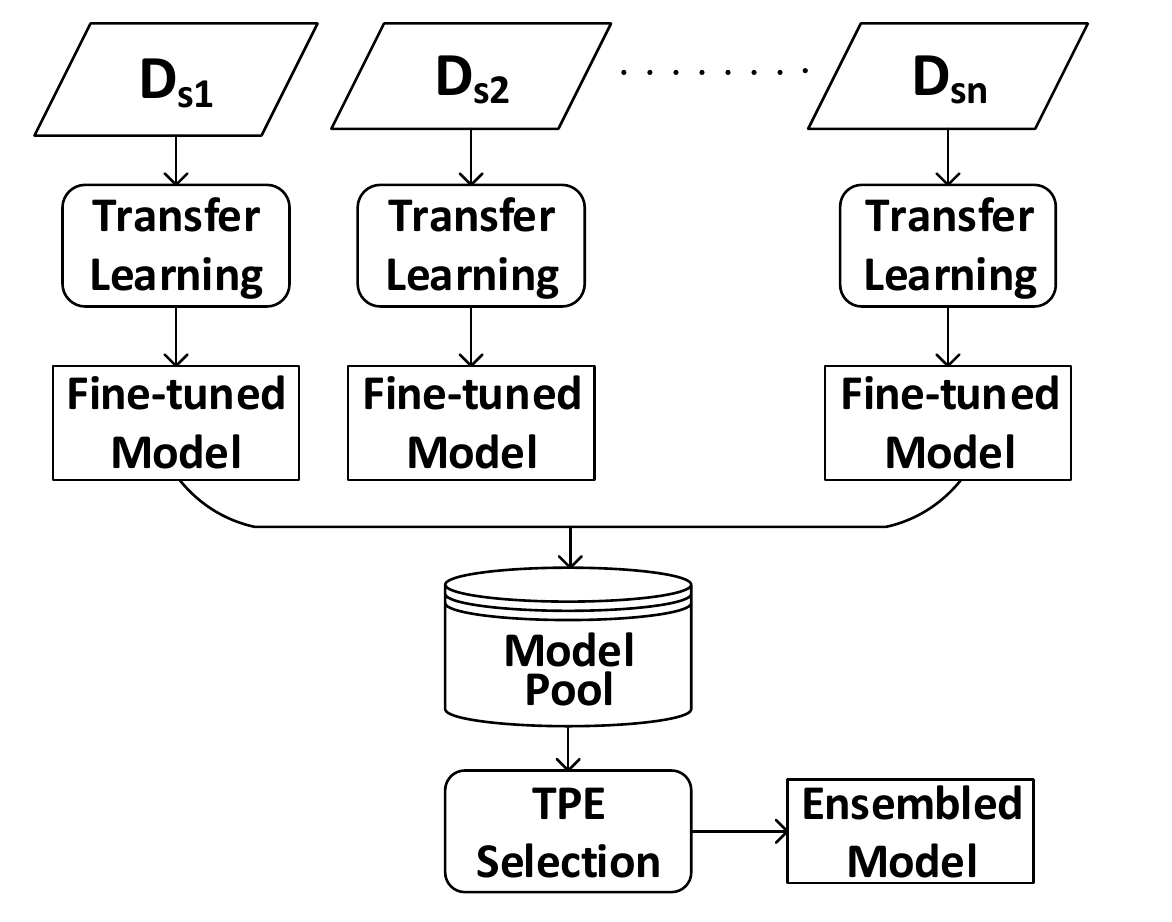}\\
	\caption{Tree-structured Parzen Estimator Ensemble Selection}
	\label{fig:tpees}
\end{figure}

In~\cite{christodoulidis2016multisource}, forward ensemble selection (FES) was used to select fine-tuned Convolution Neural Network (CNN) models from model pool. 
However,  FES is primarily designed to select models from thousands of models in~\cite{caruana2004ensemble}. 
Moreover, models with poor performance may not be selected by FES from the model pool. In such cases, it is likely to cause overfitting. To alleviate this problem, in this paper, we apply Tree-structured Parzen Estimator (TPE) to ensemble selection for multi-source transfer learning. The process of proposed Tree-structured Parzen Estimator Ensemble Selection (TPEES) is shown in Figure \ref{fig:tpees}. TPE is widely used in hyper-parameter optimization~\cite{Bergstra}. 

In Figure \ref{fig:tpees}, we use each source dataset to pre-train deep learning models. Next, the pre-trained models are fine-tuned by the target dataset. These fine-tuned models are then stored in the model pool. Afterwards, we adopt TPE algorithm in ensemble selection. We define a configuration space by setting a parameter ($\lambda^{(i)}$) for each model in model pool. In each selection iteration, TPE returns the candidate parameters $\lambda = \left\{\lambda^{(i)}\right\}_{i=1}^{n}$ with the highest Expected Improvement ($EI$). The $E I_{y^{*}}(\lambda)$ is formulated as Equation \ref{equation: tpe1} and \ref{equation: tpe2}. 
\begin{equation}\small
p(\lambda | y)=\left\{\begin{array}{ll}{\ell(\lambda)} & {\text { if } y<y^{*}} \\ {g(\lambda)} & {\text { if } y \geq y^{*}}\end{array}\right. ,
\label{equation: tpe1}
\end{equation}
where $\ell(\lambda)$ is the density formed by using the parameters $\lambda^{(i)}$ so that the corresponding loss $y^i$ is less than $y^*$, and $g(\lambda)$ is the density formed by using the remaining parameters. $y^*$ is selected to be r-quantile of the observed $y$. By construction, $\gamma=p\left(y<y^{*}\right)$. Therefore,

\begin{equation}\small
\begin{aligned} E I_{y^{*}}(\lambda) &=\frac{\gamma y^{*} \ell(\lambda)-\ell(\lambda) \int_{-\infty}^{y^{*}} p(y) d y}{\gamma \ell(\lambda)+(1-\gamma) g(\lambda)} \\ &
\propto\left(\gamma+\frac{g(\lambda)}{\ell(\lambda)}(1-\gamma)\right)^{-1} , \end{aligned}
\label{equation: tpe2}
\end{equation}
Finally, the average of ensembled model is calculated based on the output of selected models from the model pool. When $\lambda^{(i)}$ is equal to zero,  the $i^{th}$ model is not selected. $out_i$ is output of the $i^{th}$ model in the model pool. The output ($out$) of ensembled model can be formulated as Equation \ref{equation: tpees}.
\begin{equation}\small
out = (\lambda^{(1)}out_1  + \dots + \lambda^{(n)}out_n) / (\lambda^{(1)}+\dots+\lambda^{(n)}) .
\label{equation: tpees}
\end{equation}

\section{Experiments}
\begin{figure}[h]
	\centering   
	\subfigure[MLP] 
	{
		\begin{minipage}{0.2\textwidth}
			\centering          
			\includegraphics[scale=0.5]{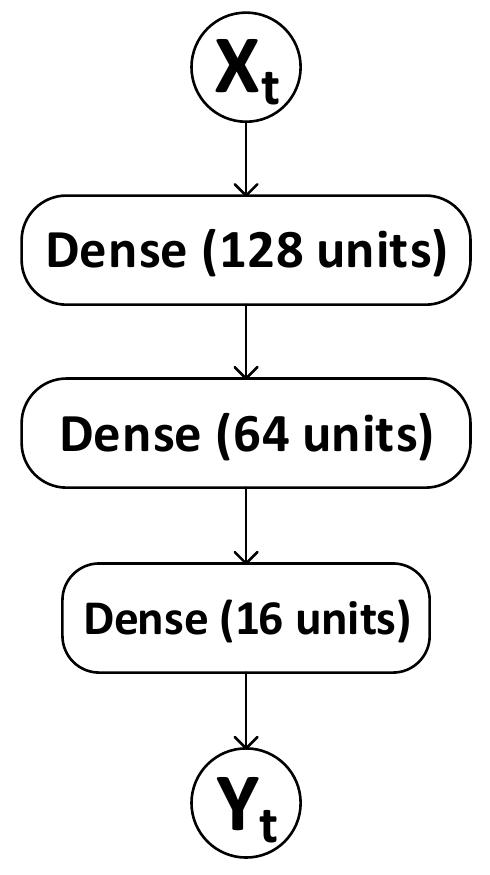} 
			\label{fig:mlp}
		\end{minipage}
	}
	\subfigure[LSTM] 
	{
		\begin{minipage}{0.2\textwidth}
			\centering   
			\includegraphics[scale=0.5]{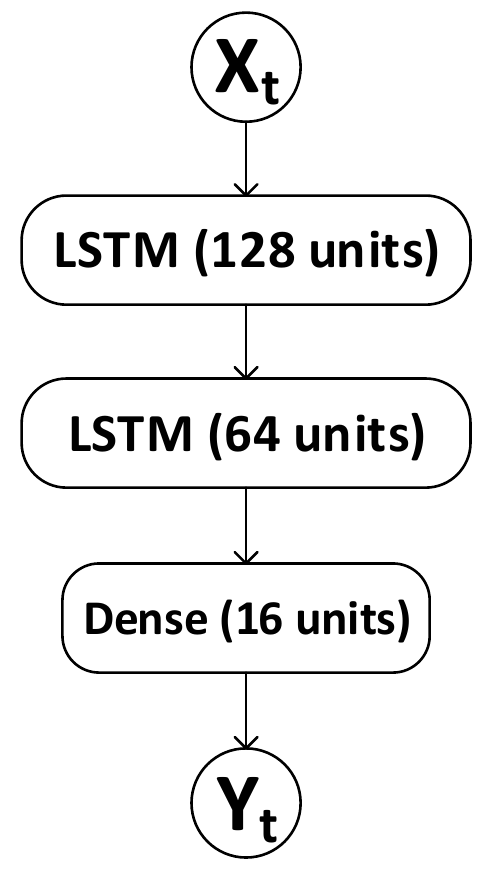}  
			\label{fig:lstm}
		\end{minipage}
	}
	\caption{MLP and LSTM architectures}
	\label{fig:architectures}  
\end{figure}

During the experiments, the proposed architecture was implemented using open source deep learning library Keras \cite{Keras} with the Tensorflow \cite{Tensorflow} back-end. The experiments were executed on Icosa Core Intel(R) Xeon (R) E5-2670 CPU @ 2.50 GHz. In order to focus on the transfer learning aspect and minimize the model's architecture and parameters involved, the same LSTM architecture used in~\cite{Roondiwala} was adopted for the experiments. The LSTM architecture is composed of a sequential input layer followed by two LSTM layers. The LSTM layers have 128 and 64 units with Tanh activation. A dense layer contains 16 units with ReLU activation and then finally a output layer with linear activation function. Besides, a Multi Layer Perceptron (MLP) model was designed based on the architecture of LSTM. The architectures of MLP and LSTM are shown in Figure \ref{fig:mlp} and \ref{fig:lstm}. Both LSTM and MLP are used in all experiments in this paper. We trained LSTM and MLP model using 22 days (trading days in majority of the stock markets are from Monday to Friday) for the look-back and 1 day for the forecast horizon. Although we only predict one time point in our experiments, our methods can be extended to predict multiple time points~\cite{corizzo2019dencast} \cite{cheng2006multistep}. Deep learning models in the proposed approach can be trained using $x$ days for the look-back and $y$ days for the forecast horizon.

In this paper, existing forecasting methods are compared against the proposed methods. These methods can be divided into 4 categories. Except the first category WTL, all categories are based on transfer learning model.  


\begin{enumerate}
	\item Without Transfer Learning (WTL): Training models without transfer learning including Autoregressive Integrated Moving Average (ARIMA), Support Vector Regression (SVR), Multilayer Perceptron (MLP) and Long Short-Term Memory (LSTM).
	\item Single Best (SB): We use one source dataset to pre-train MLP and LSTM and one target dataset to fine-tune MLP and LSTM. Among the multi-source datasets, we record the best single-source transfer results in MLP and LSTM. 
	\item Multi-source MLP (MSM): We use multi-source datasets to pre-train MLP and one target dataset to fine-tune MLP. This category includes Multi-task Learning (MTL), Average Ensemble (AE), Weighted Average Ensemble for Transfer Learning (WAETL), Forward Ensemble Selection (FES), and Tree-structured Parzen Estimator Ensemble Selection (TPEES).
	\item Multi-source LSTM (MSL): We use multi-source datasets to pre-train LSTM and one target dataset to fine-tune LSTM. This category includes Multi-task Learning (MTL), Average Ensemble (AE), Weighted Average Ensemble for Transfer Learning (WAETL), Forward Ensemble Selection (FES), and Tree-structured Parzen Estimator Ensemble Selection (TPEES).
\end{enumerate}

\noindent \paragraph{Training without Transfer Learning:}
For training without transfer learning, we use Bayesian optimization to select hyper-parameters of LSTM and MLP model and use gird search to choose hyper-parameters of ARIMA and SVR. For LSTM and MLP, we search hyper-parameters including the number of epochs ($E$), learning rate ($\alpha$), the size of mini-batch ($B$) and optimizer ($O$) within the ranges of [100-2000], [0-0.001], [16, 64, 128, 256, 512, 1024] and [Adam, SGD, RMSProp], respectively. Adaptive Moment Estimation (Adam), Stochastic Gradient Descent (SGD), RMSprop are gradient descent optimization algorithms. The loss function used in the experiment is Mean Square Error (MSE).

\noindent \paragraph{Training with Transfer Learning:}
For training with transfer learning, we use the LSTM and MLP architecture of Roondiwala et al.~\cite{Roondiwala} for forecasting. The hyper-parameters of $E$, $\alpha$, $B$ and $O$ are chosen via Bayesian optimization within the ranges of [100-1000], [0.001-0.00001], [16, 64, 128, 256, 512, 1024] and [Adam, SGD, RMSProp], respectively. The loss function used in the experiment is Mean Square Error (MSE).

\subsection{Datasets}
Datasets used in the experiments are downloaded from Yahoo Finance (\url{https://finance.yahoo.com/}).
Three different groups of datasets are selected. They are listed in Table \ref{tab:Datasets}. G1 is the  stocks of banks in Hang Seng Index (HSI) which includes HSBC, HSB, CCB, BOCHK, BOCOM and BOC as source datasets and ICBC as target dataset.
G2 is the health related stocks which includes MRK, NVS, PFE and UNH as source datasets and JNJ as target dataset. G3 is the energy related stocks which includes CVX, RDS-B, TOT and XOM as source datasets and PTR as target dataset. The range of all stocks from the datasets are from 2015 to 2019.

\begin{table}[!hbt]		
	\centering
	\scalebox{0.85}{
		\begin{tabular}{lll}
			\hline
			Group & Full Name & Short Name\\
			\hline
			
			&The Hongkong and Shanghai Banking Corporation & HSBC\\
			&Hang Seng Bank Limited & HSB\\
			G1   &China Construction Bank Corporation & CCB\\ 
			&Bank of China (Hong Kong) Limited & BOCHK\\
			&Bank of Communications Co., Ltd & BOCOM\\
			&Bank of China Limited & BOC\\
			&\textbf{Industrial and Commercial Bank of China} & \textbf{ICBC}\\
			\cline{1-3}   
			&Merck \& Co., Inc.& MRK \\
			G2    &Novartis AG & NVS\\
			&Pfizer Inc. & PFE\\
			&UnitedHealth Group Incorporated& UNH\\
			&\textbf{Johnson} \& \textbf{Johnson}& \textbf{JNJ}\\
			\cline{1-3}   
			&Chevron Corporation& CVX\\
			G3    &Royal Dutch Shell PLC & RDS-B\\
			&TOTAL S.A. & TOT\\
			&Exxon Mobil Corporation& XOM\\
			&\textbf{PetroChina Company Limited}& \textbf{PTR} \\
			\cline{1-3}

			\hline
	\end{tabular}}
	\caption{Datasets Used in the Experiments}
	\label{tab:Datasets}
\end{table}

In the experiments, time series data have been preprocessed before they are fed into supervised learning model. First, time series dataset are transformed into acceptable dataset format. The input vector $x$ consists of 22-day historical close price of stock: $x = [p_{(t)},\dots, p_{(t-21)}]$ and the output vector $y$ consists of 1-day stock price from time $t$: $y = p_{t+1}$. We use min-max scaler to rescale the time series data in [-1, 1] interval. In the experiments, we used 60\%, 20\% and 20\% of the target dataset for training, validating and testing.
After the learning process, the output of the model are inverse-normalized before computing the indicators. In this paper, we choose three classical indicators ($MAPE$, $RMSE$ and $R^2$) to measure the predictive accuracy of each model. 
$MAPE$ measures the size of the error. $RMSE$ is the mean of the square root of the error between the predicted value and the true value. $R^2$ is used for evaluating the fitting situation of the prediction model. The lower the $MAPE$ and $RMSE$, the better the model in forecasting. In contrast, higher the $R^2$, better the trained model.

\subsection{Error comparison}
In the experiments, we compare our proposed multi-source transfer learning WAETL and TPEES with other different forecasting methods. The experiment results are listed in Table \ref{tab:result1}. The proposed methods are listed in bold letters. 

\begin{table}[!hbt]		
	\centering
	\scalebox{0.85}{
		\begin{tabular}{llllll}
			\hline
			Group & Category & Model & MAPE & RMSE &  R$^2$\\

			\hline
			&         &ARIMA    &4.7079	&0.3300	&-0.2965     \\
			&Without Transfer         &SVR      &0.9739	&0.0746	&0.9347     \\
			&Learning (WTL)         &MLP      &1.0326	&0.0783	&0.9282     \\
			&         &LSTM     &0.9499	&0.0738	&0.9362  \\
			\cline{2-6}
			&Single         &MLP   &0.9495	&0.0733	&0.9371\\
			&Best (SB)        &LSTM    &0.9059	&0.0707	&0.9416\\
			\cline{2-6}
			&         &MTL   &1.0027	&0.0767	&0.9312\\
			&Multi-        &AE      &0.9448	&0.0732	&0.9373     \\
			G1          &source        &\textbf{WAETL}     &0.9297	&0.0720	&0.9394    \\
			&MLP (MSM)        &FES      &0.9282	&0.0718	&0.9398     \\
			&          &\textbf{TPEES}    &0.9186	&0.0715	   &0.9402     \\
			\cline{2-6}
			&         &MTL  & 0.9375	&0.0734	&0.9371\\
			&Multi         &AE      &0.8977	&0.0710	&0.9410     \\
			&source         &\textbf{WAETL}    &0.8962	&0.0710	&0.9410     \\
			&LSTM (MSL)        &FES     &0.8980	&0.0710	&0.9411     \\
			&         &\textbf{TPEES}  &\textbf{0.8888}	&\textbf{0.0705}	& \textbf{0.9419}     \\
			
			
			\hline
			
			&         &ARIMA    &6.4239	&10.4424 &-2.9508  \\
			&Without Transfer         &SVR      &1.1149	&2.1166	&0.8381    \\
			&Learning (WTL)          &MLP      &0.9268	&1.8342	&0.8774     \\
			&      &LSTM     &0.8201	&1.7115	&0.8932     \\
			\cline{2-6}
			&Single         &MLP   &0.8401	&1.6858	&0.8964\\
			&Best (SB)        &LSTM    &0.8176	&1.7116	&0.8932\\
			\cline{2-6}
			
			&         &MTL  &0.8447	&1.7452	&0.8890\\
			&Multi-         &AE     &0.8168	&1.6862	&0.8964\\
			G2          &source       &\textbf{WAETL}    &0.8663	&1.7470	&0.8888\\
			&MLP (MSM)          &FES     &0.8109	&1.6713	&0.8982\\
			&         &\textbf{TPEES}   &\textbf{0.8083}	&\textbf{1.6693}	&\textbf{0.8984}\\
			\cline{2-6}
			&        &MTL    &1.5821	&2.7568	&0.7230\\
			&Multi-         &AE      &0.8892	&1.7220	&0.8919 \\
			&source         &\textbf{WAETL}    &0.8637	&1.7064	&0.8939 \\
			&LSTM (MSL)         &FES     &0.8488	&1.6984	&0.8949 \\
			&         &\textbf{TPEES}   &0.8373	&1.6930	&0.8955\\
			\hline

			&         &ARIMA    &18.7188 &11.4203 &-0.9829 \\
			&Without Transfer        &SVR      &3.5668	&2.6927	&0.8853     \\
			&Learning (WTL)         &MLP      &2.6026	&1.8314	&0.9463    \\
			&     &LSTM     &1.4969	&1.1651	&0.9783    \\
			\cline{2-6}
			&Single         &MLP   &\textbf{1.4233}	&\textbf{1.1215}	&\textbf{0.9799}\\
			&Best (SB)        &LSTM &1.6093	&1.2128	&0.9764\\
			
			\cline{2-6}
			&         &MTL  &1.7025	&1.2763	&0.9739\\
			&Multi-        &AE      &1.6940	&1.2672	&0.9743\\
			G3          &source        &\textbf{WAETL}    &1.8794	&1.3879	&0.9691\\
			&MLP          &FES     &1.5585	&1.1895	&0.9773\\
			&(MSM)         &\textbf{TPEES}    &1.5217	&1.1692	&0.9781 \\ 
			\cline{2-6}
			&         &MTL  &2.2074	&1.6122	&0.9584\\
			&Multi-         &AE      &1.7058	&1.2777	&0.9738 \\
			&source         &\textbf{WAETL}    &1.7620	&1.3234	&0.9719\\
			&LSTM (MSL)         &FES     &1.7114	&1.2807	&0.9737 \\
			&         &\textbf{TPEES}  &1.7042	&1.2754	&0.9739\\
			\hline
	\end{tabular}}
	\caption{Experiment Results for Different Forecasting Methods}
	\label{tab:result1}
\end{table}

From these results, we can observe that ARIMA is not suitable for financial time series forecasting because it always obtains the worst performance. We can also observe that the performance of MLP and LSTM are better than ARIMA and SVR in most of the cases. Besides, models with transfer learning have significant impact on time series forecasting in most of cases. However, in some situation, we can find that the results of LSTMs in Single Best (SB) and Multi-source category are worse than LSTM in Without Transfer Learning (WTL) category. This situation is often labeled as negative transfer learning \cite{Rosenstein}. In G1 and G2, we can also observe that results of models in multi-source category are better than models from single best category. In addition, we can observe that results of WAETL are worse than results of TPEES. It is possible that the performance of some models in the model pool is indeed poor, resulting in a inferior WAETL model. Furthermore, the similarity of source and target dataset calculated by the distance function may not be accurate. However, in TPEES, poor models in model pool may be not selected since TPEES model selection is based on their impact on the ensembled model. All in all, TPEES achieves the best results in majority of the cases.

To further investigate the performance of the proposed approaches, we conduct more detailed experiments on dataset G1. In this experiment, we use each stock in G1 as a target dataset and the rest of the stocks as source datasets. The experiment results are listed in Table \ref{tab:HSBC}, \ref{tab:HSB}, \ref{tab:CCB}, \ref{tab:BOCHK}, \ref{tab:BOC} and \ref{tab:BOCOM}. From these results, we found that models in multi-source MLP, multi-source LSTM and single best category are better than models from Without Transfer Learning category in most of the cases. In Table \ref{tab:HSBC} (HSBC), Table \ref{tab:BOCHK} (BOCHK), Table \ref{tab:BOCOM} (BOCOM), Table \ref{tab:BOC} (BOC), the performance of MTLs in multi-source (MLP) is better than in multi-source LSTM. Besides, the best results are found in multi-source MLP and multi-source LSTM category. TPEES also achieves best results in Table \ref{tab:HSBC} (HSBC), Table \ref{tab:CCB} (CCB), Table \ref{tab:BOCHK} (BOCHK), and Table \ref{tab:BOC} (BOC).
  
\begin{table}[!hbt]		
	\begin{center}
		\scalebox{0.9}{
			\begin{tabular}{lllll}
				\hline
				Category &Model & MAPE & RMSE &  R$^2$\\
				\hline
				&ARIMA      &3.9042	&3.0650	&-0.7349 \\
				Without Transfer   &  SVR   &0.8782	&0.7509	&0.8945     \\
				Learning (WTL)   &  MLP     &0.8264	&0.7260	&0.8999    \\
				&LSTM    &0.7481	&0.6649	&0.9160     \\
				\cline{1-5}
				Single         &MLP   &0.7737	&0.6801	&0.9121\\
				Best (SB)        &LSTM &0.7335	&0.6552	&0.9185\\
				\cline{1-5}
				&MTL   &0.7476	&0.6568	&0.9181\\
				Multi-    &AE    &0.7480	&0.6565	&0.9181     \\ 
				source    &\textbf{WAETL}     &0.7554	&0.6633	&0.9164 \\
				MLP (MSM)     &FES       &0.7516	&0.6598	&0.9173 \\
				&\textbf{TPEES}      &0.7486	&0.6554	&0.9184     \\
				\cline{1-5}
				&MTL  &0.7930	&0.6959	&0.9080\\
				Multi-     &AE     &0.7502	&0.6654	&0.9159 \\
				source     &\textbf{WAETL}    & 0.7348	&0.6567	&0.9181 \\
				LSTM (MSL)     &FES     &0.7509	&0.6666	&0.9156 \\
				&\textbf{TPEES}     &\textbf{0.7307}	&\textbf{0.6536}	&\textbf{0.9189}    \\
				
				\hline
		\end{tabular}}
	\end{center}
	\caption{Experiment Results for HSBC as Target Dataset and HSB, CCB, BOCHK, BOCOM, BOC, ICBC as Source Datasets}
	\label{tab:HSBC}
\end{table}

\begin{table}[!hbt]		
	\centering
	\scalebox{0.9}{
		\begin{tabular}{lllll}
			\hline
			Category &Model & MAPE & RMSE &  R$^2$\\
			\hline
			&ARIMA      &14.9583	&30.5365	&-5.6414 \\
			Without Transfer     &SVR    &1.3342	&3.2411	&0.9236     \\
			Learning (WTL)     &MLP     &1.3209	&3.2278	&0.9235    \\
			&LSTM     &1.2924	&3.2787	&0.9211     \\
			\cline{1-5}   
			Single         &MLP   &0.9053	&2.3615	&0.9591\\
			Best (SB)        &LSTM &0.8928	&2.3208	&0.9605\\
			\cline{1-5}
			&MTL  &0.9335	&2.3783	&0.9585\\
			Multi-     &AE    &0.9250	&2.3707	&0.9587    \\
			source     &\textbf{WAETL}    &0.8922	&2.3408	&0.9598 \\
			MLP (MSM)     &FES      &0.8910	&\textbf{2.3061}	&\textbf{0.9610} \\
			&\textbf{TPEES}    &0.8970	&2.3165	&0.9606     \\
			\cline{1-5}
			&MTL &0.9177	&2.3681	&0.9588\\
			Multi-     &AE     & 0.8856	&2.3079	&0.9609 \\
			source     &\textbf{WAETL}    &0.9090	&2.3551	&0.9593 \\
			LSTM (MSL)     &FES     &0.8879	&2.3221	&0.9604 \\
			&\textbf{TPEES}   & \textbf{0.8854}	&2.3157	&0.9606    \\
			
			\hline
	\end{tabular}}
	\caption{Experiment Results for HSB as Target Dataset and HSBC, CCB, BOCHK, BOCOM, BOC, ICBC as Source Datasets}
	\label{tab:HSB}
\end{table}

\begin{table}[!hbt]		
	\centering
	\scalebox{0.9}{
		\begin{tabular}{lllll}
			\hline
			Category &Model & MAPE & RMSE &  R$^2$\\
			\hline
			&ARIMA      &4.9897	&0.3834	&-0.1024 \\   
			Without Transfer     &SVR    &0.9828	&0.0856	&0.9459    \\
			Learning (WTL)     &MLP     &1.0673	&0.0917	&0.9379    \\
			&LSTM     &\textbf{0.9026}	&0.0826	&0.9497     \\
			\cline{1-5}   
			Single         &MLP   &0.9294	&0.0837	&0.9483\\
			Best (SB)         &LSTM &0.9089	&0.0824	&0.9498\\
			\cline{1-5}
			&MTL &1.0564	&0.0898	&0.9404 \\
			Multi-     &AE   &0.9280	&0.0828	&0.9494    \\
			source     &\textbf{WAETL}     &0.9509	&0.0844	&0.9474 \\
			MLP (MSM)    &FES      &0.9201	&0.0836	&0.9483 \\
			&\textbf{TPEES}   & 0.9115	&0.0827	&0.9495     \\
			\cline{1-5}
			&MTL &0.9405	&0.0839	&0.9481\\
			Multi-     &AE     &0.9175	&0.0824	&0.9499 \\
			source     &\textbf{WAETL}     &0.9177	&0.0823	&0.9500 \\
			LSTM (MSL)    &FES     &0.9151	&0.0820	&\textbf{0.9504} \\
			&\textbf{TPEES}  &0.9157	&\textbf{0.0819}	&\textbf{0.9504}    \\
			\hline
	\end{tabular}}
	\caption{Experiment Results for CCB as Target Dataset and HSBC, HSB, BOCHK, BOCOM, BOC, ICBC as Source Datasets}
	\label{tab:CCB}
\end{table}

\begin{table}[!hbt]		
	\centering
	\scalebox{0.9}{
		\begin{tabular}{lllll}
			\hline
			Category &Model & MAPE & RMSE &  R$^2$\\
			\hline
			&ARIMA       &21.2078	&7.0005	&-7.1548 \\
			Without Transfer     &SVR    &1.5240	&0.6362	&0.9298   \\
			Learning (WTL)     &MLP     &1.4428	&0.5976	&0.9367    \\
			&LSTM    &1.2458	&0.5461	&0.9471     \\
			\cline{1-5}   
			Single         &MLP  &1.1838	&0.4889	&0.9576\\
			Best (SB)        &LSTM &1.0467	&0.4632	&0.9619\\
			\cline{1-5}
			&MTL &1.2201	&0.5309	&0.9500 \\
			Multi-     &AE   &1.0866	&0.4788	&0.9593   \\
			source     &\textbf{WAETL}    &1.0822	&0.4757	&0.9599 \\
			MLP (MSM)     &FES      &1.0708	&0.4759	&0.9598 \\
			&\textbf{TPEES}   &1.0612	&0.4700	&0.9608    \\
			\cline{1-5}
			&MTL &1.7555	&0.7371	&0.9036\\
			Multi-     &AE    &1.0551	&0.4675	&0.9612 \\
			source     &\textbf{WAETL}    &\textbf{1.0298}	&0.4673	&0.9613 \\
			LSTM (MSL)     &FES   &1.0440	&0.4618	&0.9622 \\
			&\textbf{TPEES} &1.0369	&\textbf{0.4613}	&\textbf{0.9623}   \\
			\hline
	\end{tabular}}
	\caption{Experiment Results for BOCHK as Target Dataset and HSBC, HSB, CCB, BOCOM, BOC, ICBC as Source Datasets}
	\label{tab:BOCHK}
\end{table}

\begin{table}[!hbt]		
	\centering
	\scalebox{0.9}{
		\begin{tabular}{lllll}
			\hline
			Standards &Model & MAPE & RMSE &  R$^2$\\
			\hline
			
			&ARIMA       &5.7815	&0.2410	&-0.4019 \\
			Without Transfer     &SVR    &0.9247	&0.0435	&0.9551  \\
			Learning (WTL)     &MLP     &0.8520	&0.0415	&0.9593    \\
			&LSTM     &0.7812	&0.0397	&0.9628     \\
			\cline{1-5}   
			Single         &MLP  &0.8629	&0.0409	&0.9604\\
			Best (SB)        &LSTM &0.7825	&0.0395	&0.9632\\
			\cline{1-5}
			&MTL  &0.8393	&0.0410	&0.9602\\
			Multi-     &AE   &0.8612	&0.0412	&0.9599   \\
			source     &\textbf{WAETL}  &0.8483	&0.0406	&0.9611\\
			MLP (MSM)     &FES     &0.8358&0.0405	&0.9612 \\
			&\textbf{TPEES}   &0.8349	&0.0404	&0.9614    \\
			\cline{1-5}
			&MTL &2.3387	&0.1085	&0.7218 \\
			Multi-     &AE   &0.7795	&0.0394	&0.9633 \\ 
			source     &\textbf{WAETL}    &0.7866	&0.0395	&0.9632 \\
			LSTM (MSL)     &FES    &\textbf{0.7759}	&0.0394	&0.9632 \\
			&\textbf{TPEES} & 0.7766	&\textbf{0.0393}	&\textbf{0.9634}  \\
			\hline
	\end{tabular}}
	\caption{Experiment Results for BOC as Target Dataset and HSBC, HSB, CCB, BOCHK, BOCOM, ICBC as Source Datasets}
	\label{tab:BOC}
\end{table}

\begin{table}[!hbt]		
	\centering
	\scalebox{0.9}{
		\begin{tabular}{lllll}
			\hline
			Category &Model & MAPE & RMSE &  R$^2$\\
			\hline
			&ARIMA       &10.4469	&0.7643	&-2.5990 \\
			Without Transfer     &SVR    & 0.9039	&0.0808	&0.9597   \\
			Learning (WTL)     &MLP     &0.9998	&0.0860	&0.9545    \\
			&LSTM     & 0.9684	&0.0856	&0.9549     \\
			\cline{1-5}   
			Single         &MLP  &0.9640	&0.0819	&0.9587\\
			Best (SB)        &LSTM &\textbf{0.9024}	&0.0798	&0.9608\\
			\cline{1-5}
			&MTL &0.9668	&0.0830	&0.9576 \\
			Multi-     &AE   &0.9198	&0.0803	&0.9603    \\
			source     &\textbf{WAETL}   &0.9885	&0.0846	&0.9559 \\
			MLP (MSM)     &FES      &0.9265	&0.0806	&0.9600 \\
			&\textbf{TPEES}   &0.9449	&0.0817	&0.9589     \\
			\cline{1-5}
			&MTL &1.3794	&0.1196	&0.9119\\
			Multi-     &AE    &0.9087	&\textbf{0.0797}	&\textbf{0.9609} \\
			source     &\textbf{WAETL}   &0.9525	&0.0822	&0.9584 \\
			LSTM (MSL)     &FES    &0.9218	&0.0804	&0.9602 \\
			&\textbf{TPEES} & 0.9116	&\textbf{0.0797}	&0.9608   \\
			\hline
	\end{tabular}}
	\caption{Experiment Results for BOCOM as Target Dataset and HSBC, HSB, CCB, BOCHK, BOC, ICBC as Source Datasets}
	\label{tab:BOCOM}
\end{table}

\subsection{Evaluation of distance functions}
In this paper, we proposed two ensemble methods for multi-source transfer learning. TPEES selects models based on the performance of models and does not calculate the similarity between source and target datasets. However, in WAETL, the similarity between source and target datasets is used as weight. Therefore, in addition to the error comparison, we further investigate the performance of different distance functions which are used for calculating the weights in WAETL. In this experiment, we compare the result of WAETL when different distance functions are used. These algorithms include CORrelation ALignment (CORAL) loss \cite{sun2016return}, Wasserstein Distance (WD) \cite{Ruschendorf1985Wasserstein}, Dynamic Time Warping (DTW)\cite{Berndt1994Dynamic}, and Pearson Correlation Coefficient (PCC)\cite{benesty2009pearson}. The results are listed in Table \ref{tab:distanceMAPE}, \ref{tab:distanceRMSE}, and \ref{tab:distanceR2}. 

\begin{table}[!hbt]		
	\begin{center}	
		\scalebox{0.9}{		
			\begin{tabular}{c|c|c|c|c}		
				\hline		
				\multirow{2}{*}{$D_t$}&                  		
				\multicolumn{4}{|c}{WAETL MLP} \\		
				\cline{2-5} 		
				& Coral &WD &DTW & PCC\\		
				\hline		
				HSBC&0.7554&\textbf{0.7434}&0.7497&0.7485 \\	
				HSB&\textbf{0.8922}&0.9128&0.9046&0.9080  \\
				CCB&0.9509&0.9293&\textbf{0.9175}&0.9239 \\
				ICBC&\textbf{0.9298}&0.9499&0.9368&0.9417 \\
				BOCHK&1.0822&\textbf{1.0701}&1.0741&1.0715 \\
				BOCOM&0.9885&\textbf{0.9194}&0.9404&0.9367 \\
				BOC&0.8483&0.8371&\textbf{0.8353}&0.8354 \\
				\hline
				\multirow{2}{*}{$D_t$}&  \multicolumn{4}{|c}{WAETL LSTM}\\	
				\cline{2-5} 		
				& Coral &WD &DTW & PCC\\	
				\hline	
				HSBC &\textbf{0.7348}&0.7410&0.7415&0.7484\\	
				HSB  &0.9090&\textbf{0.8854}&0.8862&0.8864\\
				CCB  &0.9177&0.9163&\textbf{0.9146}&0.9161\\
				ICBC  &0.8962&\textbf{0.8945}&0.8950&0.8969\\
				BOCHK &\textbf{1.0298}&1.0413&1.0428&1.0427\\
				BOCOM  &0.9525&0.9155&0.9139&\textbf{0.9117}\\
				BOC &0.7866&\textbf{0.7776}&0.7811&0.7825 \\
				\hline	
		\end{tabular}}		
	\end{center}		
	\caption{$MAPE$ of Different Distance Functions Used in WAETL.}		
	\label{tab:distanceMAPE}	
\end{table}

\begin{table}[!hbt]		
	\begin{center}
		\scalebox{0.9}{		
			\begin{tabular}{c|c|c|c|c}			
				\hline		
				\multirow{2}{*}{$D_t$}&                  		
				\multicolumn{4}{|c}{WAETL MLP} \\		
				\cline{2-5} 		
				& Coral &WD &DTW & PCC\\			
				\hline		
				HSBC&0.6633&0.6576&0.6593&\textbf{0.6569}   \\
				HSB&\textbf{2.3408}&2.3477&2.3416&2.3454    \\
				CCB&0.0844&0.0836&0.0828&\textbf{0.0827}     \\
				ICBC&\textbf{0.0720}&0.0736&0.0728&0.0730   \\
				BOCHK&0.4757&0.4727&0.4724&\textbf{0.4723}  \\
				BOCOM&0.0846&\textbf{0.0806}&0.0814&0.0812  \\
				BOC&0.0406&0.0406&0.0405&\textbf{0.0405}     \\
				\hline
				\multirow{2}{*}{$D_t$}&  \multicolumn{4}{|c}{WAETL LSTM}\\	
				\cline{2-5} 		
				& Coral &WD &DTW & PCC\\	
				\hline	
				HSBC &\textbf{0.6567}&0.6598&0.6606&0.6645\\
				HSB  &2.3551&\textbf{2.3086}&2.3091&2.3097\\
				CCB  &\textbf{0.0823}&0.0825&0.0825&0.0824\\
				ICBC  &0.0710&\textbf{0.0709}&0.0709&0.0710\\
				BOCHK  &0.4673&\textbf{0.4637}&0.4644&0.4648\\
				BOCOM  &0.0822&0.0801&0.0802&\textbf{0.0801}\\
				BOC   &0.0395&\textbf{0.0393}&0.0394&0.0394 \\
				\hline	
		\end{tabular}}		
	\end{center}		
	\caption{$RMSE$ of Different Distance Functions Used in WAETL.}		
	\label{tab:distanceRMSE}	
\end{table}

\begin{table}[!hbt]		
	\begin{center}
		\scalebox{0.9}{		
			\begin{tabular}{c|c|c|c|c}			
				\hline		
				\multirow{2}{*}{$D_t$}&                  		
				\multicolumn{4}{|c}{WAETL MLP} \\		
				\cline{2-5} 		
				& Coral &WD &DTW & PCC\\		
				\hline		
				HSBC& 0.9164&0.9179&0.9174&\textbf{0.9180}  \\
				HSB&\textbf{0.9598}&0.9595&0.9597&0.9596     \\
				CCB&0.9474&0.9483&\textbf{0.9495}&0.9494   \\
				ICBC&\textbf{0.9394}&0.9367&0.9381&0.9377    \\
				BOCHK&0.9599&0.9604&0.9604&\textbf{0.9605}   \\
				BOCOM&0.9559&\textbf{0.9600}&0.9592&0.9594  \\
				BOC&0.9611&0.9611&0.9611&\textbf{0.9612}    \\
				\hline
				\multirow{2}{*}{$D_t$}&  \multicolumn{4}{|c}{WAETL LSTM}\\	
				\cline{2-5} 		
				& Coral &WD &DTW & PCC\\	
				\hline	
				HSBC &\textbf{0.9181}&0.9173&0.9171&0.9161\\
				HSB &0.9593&\textbf{0.9609}&0.9609&0.9608\\
				CCB &\textbf{0.9500}&0.9497&0.9498&0.9499\\
				ICBC &0.9410&0.9412&\textbf{0.9413}&0.9411\\
				BOCHK &0.9613&\textbf{0.9619}&0.9618&0.9617\\
				BOCOM  &0.9584&\textbf{0.9605}&0.9604&0.9605\\
				BOC  &0.9632&\textbf{0.9633}&0.9632&0.9632\\
				\hline
		\end{tabular}}		
	\end{center}		
	\caption{$R^2$ of Different Distance Functions Used in WAETL.}		
	\label{tab:distanceR2}	
\end{table}

From the MAPE results (Table \ref{tab:distanceMAPE}), we can observe that Coral obtains the best results four times, WD achieves six times, DTW achieves three times and PCC achieves once. From the $RMSE$ results (Table \ref{tab:distanceRMSE}), Coral obtains the best results four times, WD achieves five times, DTW achieves none, and PCC achieves five times. From the $R^2$ results (Table \ref{tab:distanceR2}), Coral obtains the best results four times, WD achieves five times, DTW achieves two times and PCC achieves three times. Although we find that WD and Coral do not always produce the best results, they are the most stable and robust among all the tested functions. Therefore, we can conclude that utilizing WD and Coral to calculate the weights for WAETL can get lower $MAPE$, lower $RMSE$ and higher $R^2$ in time series forecasting.

\section{Conclusion}
In this paper, we propose two multi-source transfer learning methods namely Weighted Average Ensemble for Transfer Learning (WAETL) and Tree-structured Parzen Estimator Ensemble Selection (TPEES). Extensive experiments are conducted to compare the performance of the proposed approaches with other competing methods. The experiment results reveal that TPEES achieves best result in most of the cases. In addition, we further analyze the impact of four similarity functions for multi-source transfer learning. We found that WD and Coral distance functions achieve favorable results when they are used for calculating the weights in WAETL approach. The main contributions of this paper are as follows. First, the proposed approaches allow the effective use of multiple source datasets for training in financial time series forecasting. In other words, the proposed approaches effectively solve the insufficient training data problem in developing deep learning models for financial domain. Second, our approach demonstrates that multi-source transfer learning can be applied to exploit the correlation among stocks from the same industry. Third, our evaluation on using different distance functions can be used as a guideline for calculating the distance among sources in instance based multi-source transfer learning. 
As for the future work, we are planning to extend our models to take into account negative correlation and other technical indicators from stock market data. 




%
%
%
\bibliographystyle{IEEEtran}
\bibliography{reference}

\begin{thebibliography}{10}
\providecommand{\url}[1]{#1}
\csname url@samestyle\endcsname
\providecommand{\newblock}{\relax}
\providecommand{\bibinfo}[2]{#2}
\providecommand{\BIBentrySTDinterwordspacing}{\spaceskip=0pt\relax}
\providecommand{\BIBentryALTinterwordstretchfactor}{4}
\providecommand{\BIBentryALTinterwordspacing}{\spaceskip=\fontdimen2\font plus
\BIBentryALTinterwordstretchfactor\fontdimen3\font minus
  \fontdimen4\font\relax}
\providecommand{\BIBforeignlanguage}[2]{{%
\expandafter\ifx\csname l@#1\endcsname\relax
\typeout{** WARNING: IEEEtran.bst: No hyphenation pattern has been}%
\typeout{** loaded for the language `#1'. Using the pattern for}%
\typeout{** the default language instead.}%
\else
\language=\csname l@#1\endcsname
\fi
#2}}
\providecommand{\BIBdecl}{\relax}
\BIBdecl

\bibitem{karl2016survey}
\BIBentryALTinterwordspacing
K.~Weiss, T.~M. Khoshgoftaar, and D.~Wang, ``A survey of transfer learning,''
  \emph{Journal of Big Data}, vol.~3, no.~1, p.~9, 2016. [Online]. Available:
  \url{https://doi.org/10.1186/s40537-016-0043-6}
\BIBentrySTDinterwordspacing

\bibitem{Ding}
X.~Ding, Y.~Zhang, T.~Liu, and J.~Duan, ``Deep learning for event-driven stock
  prediction,'' in \emph{Proceedings of the $24^{th}$ International Conference
  on Artificial Intelligence}.\hskip 1em plus 0.5em minus 0.4em\relax AAAI
  Press, 2015, pp. 2327--2333.

\bibitem{bao2017deep}
W.~Bao, J.~Yue, and Y.~Rao, ``A deep learning framework for financial time
  series using stacked autoencoders and long-short term memory,'' \emph{PLOS
  ONE}, vol.~12, no.~7, p. e0180944, 2017.

\bibitem{Makridakis}
S.~Makridakis, E.~Spiliotis, and V.~Assimakopoulos, ``Statistical and machine
  learning forecasting methods: Concerns and ways forward,'' \emph{PLOS ONE},
  vol.~13, no.~3, pp. 1--26, 2018.

\bibitem{Yosinski}
J.~Yosinski, J.~Clune, Y.~Bengio, and H.~Lipson, ``How transferable are
  features in deep neural networks?'' in \emph{Advances in Neural Information
  Processing Systems}, 2014, pp. 3320--3328.

\bibitem{Fawaz}
H.~I. Fawaz, G.~Forestier, J.~Weber, L.~Idoumghar, and P.-A. Muller, ``Transfer
  learning for time series classification,'' in \emph{2018 IEEE International
  Conference on Big Data (Big Data)}.\hskip 1em plus 0.5em minus 0.4em\relax
  IEEE, 2018, pp. 1367--1376.

\bibitem{Laptev}
N.~Laptev, J.~Yu, and R.~Rajagopal, ``Reconstruction and regression loss for
  time-series transfer learning,'' \emph{SIGKDD MiLeTS--2018--8c}, 2018.

\bibitem{Ye}
R.~Ye and Q.~Dai, ``A novel transfer learning framework for time series
  forecasting,'' \emph{Knowledge-Based Systems}, vol. 156, pp. 74--99, 2018.

\bibitem{Huang}
J.-T. Huang, J.~Li, D.~Yu, L.~Deng, and Y.~Gong, ``Cross-language knowledge
  transfer using multilingual deep neural network with shared hidden layers,''
  in \emph{2013 IEEE International Conference on Acoustics, Speech and Signal
  Processing}.\hskip 1em plus 0.5em minus 0.4em\relax IEEE, 2013, pp.
  7304--7308.

\bibitem{hu2016transfer}
Q.~Hu, R.~Zhang, and Y.~Zhou, ``Transfer learning for short-term wind speed
  prediction with deep neural networks,'' \emph{Renewable Energy}, vol.~85, pp.
  83--95, 2016.

\bibitem{christodoulidis2016multisource}
S.~Christodoulidis, M.~Anthimopoulos, L.~Ebner, A.~Christe, and S.~Mougiakakou,
  ``Multisource transfer learning with convolutional neural networks for lung
  pattern analysis,'' \emph{IEEE Journal of Biomedical and Health Informatics},
  vol.~21, no.~1, pp. 76--84, 2016.

\bibitem{tyagi2014survey}
V.~Tyagi and A.~Mishra, ``A survey on ensemble combination schemes of neural
  network,'' \emph{International Journal of Computer Applications}, vol.~95,
  no.~16, pp. 18--21, 2014.

\bibitem{Touretzky1997Learning}
P.~Sollich and A.~Krogh, ``Learning with ensembles: How overfitting can be
  useful,'' in \emph{Advances in Neural Information Processing Systems}, 1996,
  pp. 190--196.

\bibitem{Rosenstein}
M.~T. Rosenstein, Z.~Marx, L.~P. Kaelbling, and T.~G. Dietterich, ``To transfer
  or not to transfer,'' in \emph{NIPS 2005 Workshop on Transfer Learning}, vol.
  898, 2005, pp. 1--4.

\bibitem{mignone2019exploiting}
\BIBentryALTinterwordspacing
P.~Mignone, G.~Pio, D.~D’Elia, and M.~Ceci, ``{Exploiting transfer learning
  for the reconstruction of the human gene regulatory network},''
  \emph{Bioinformatics}, 10 2019, btz781. [Online]. Available:
  \url{https://doi.org/10.1093/bioinformatics/btz781}
\BIBentrySTDinterwordspacing

\bibitem{sun2016return}
\BIBentryALTinterwordspacing
B.~Sun, J.~Feng, and K.~Saenko, ``Return of frustratingly easy domain
  adaptation,'' in \emph{Proceedings of the Thirtieth AAAI Conference on
  Artificial Intelligence}.\hskip 1em plus 0.5em minus 0.4em\relax AAAI Press,
  2016, pp. 2058--2065. [Online]. Available:
  \url{http://dl.acm.org/citation.cfm?id=3016100.3016186}
\BIBentrySTDinterwordspacing

\bibitem{Ruschendorf1985Wasserstein}
L.~R{\"u}schendorf, ``The wasserstein distance and approximation theorems,''
  \emph{Probability Theory and Related Fields}, vol.~70, no.~1, pp. 117--129,
  1985.

\bibitem{Berndt1994Dynamic}
D.~Berndt and J.~Clifford, ``Using dynamic time warping to find patterns in
  time series,'' in \emph{KDD Workshop}, vol. 10/16, 1994, pp. 359--370.

\bibitem{benesty2009pearson}
J.~Benesty, J.~Chen, Y.~Huang, and I.~Cohen, ``Pearson correlation
  coefficient,'' in \emph{Noise Reduction in Speech Processing}.\hskip 1em plus
  0.5em minus 0.4em\relax Springer, 2009, pp. 1--4.

\bibitem{caruana2004ensemble}
R.~Caruana, A.~Niculescu-Mizil, G.~Crew, and A.~Ksikes, ``Ensemble selection
  from libraries of models,'' in \emph{Proceedings of the Twenty-first
  International Conference on Machine Learning}.\hskip 1em plus 0.5em minus
  0.4em\relax ACM, 2004, pp. 18--25.

\bibitem{Bergstra}
J.~S. Bergstra, R.~Bardenet, Y.~Bengio, and B.~K{\'e}gl, ``Algorithms for
  hyper-parameter optimization,'' in \emph{Advances in Neural Information
  Processing Systems}, 2011, pp. 2546--2554.

\bibitem{Keras}
``Keras,'' \url{https://keras.io/}, 2015.

\bibitem{Tensorflow}
M.~Abadi, P.~Barham, J.~Chen, Z.~Chen, A.~Davis, J.~Dean, M.~Devin,
  S.~Ghemawat, G.~Irving, M.~Isard \emph{et~al.}, ``Tensorflow: A system for
  large-scale machine learning,'' in \emph{$12^{th}$ {USENIX} Symposium on
  Operating Systems Design and Implementation ({OSDI} 16)}, 2016, pp. 265--283.

\bibitem{Roondiwala}
M.~Roondiwala, H.~Patel, and S.~Varma, ``Predicting stock prices using lstm,''
  \emph{International Journal of Science and Research (IJSR)}, vol.~6, no.~4,
  pp. 1754--1756, 2017.

\bibitem{corizzo2019dencast}
\BIBentryALTinterwordspacing
R.~Corizzo, G.~Pio, M.~Ceci, and D.~Malerba, ``Dencast: distributed
  density-based clustering for multi-target regression,'' \emph{Journal of Big
  Data}, vol.~6, no.~1, p.~43, Jun 2019. [Online]. Available:
  \url{https://doi.org/10.1186/s40537-019-0207-2}
\BIBentrySTDinterwordspacing

\bibitem{cheng2006multistep}
H.~Cheng, P.-N. Tan, J.~Gao, and J.~Scripps, ``Multistep-ahead time series
  prediction,'' in \emph{Advances in Knowledge Discovery and Data Mining},
  W.-K. Ng, M.~Kitsuregawa, J.~Li, and K.~Chang, Eds.\hskip 1em plus 0.5em
  minus 0.4em\relax Berlin, Heidelberg: Springer Berlin Heidelberg, 2006, pp.
  765--774.

\end{thebibliography}

\end{document}